%% file: main.tex
\documentclass[journal,twoside,web]{IEEEtran}
\makeatletter
\let\NAT@parse\undefined
\makeatother
\usepackage{tmi}
\usepackage[numbers]{natbib}
\usepackage{cite}
\usepackage{amsmath,amssymb,amsfonts}
\usepackage{layouts}
\usepackage{algorithmic}
\usepackage{graphicx, caption}
\usepackage{bm}
\usepackage{textcomp}
\usepackage{diagbox}
\usepackage{makecell}
\usepackage{hyperref}
\usepackage{multirow}
\usepackage{hhline}
\usepackage{booktabs}
\usepackage{array}
\newcolumntype{?}{!{\vrule width 1.25pt}}

\def\BibTeX{{\rm B\kern-.05em{\sc i\kern-.025em b}\kern-.08em
    T\kern-.1667em\lower.7ex\hbox{E}\kern-.125emX}}

\begin{document}
\title{Recovering high-quality FODs from a reduced number of diffusion-weighted images using a model-driven deep learning architecture}
\author{Joseph Bartlett, Catherine E. Davey, Leigh A. Johnston, \IEEEmembership{Senior Member, IEEE}, and Jinming Duan
\thanks{J. Bartlett, and J. Duan are with the School of Computer Science, the University of Birmingham, Birmingham, UK.}
\thanks{J. Duan is with the Alan Turing Institute, London, UK.}
\thanks{J. Bartlett, C. E. Davey and L. A. Johnston are with the Department of Biomedical Engineering, the Melbourne Brain Centre Imaging Unit and the Graeme Clark Institute, the University of Melbourne, Melbourne, Australia.}
\thanks{The corresponding author is J. Duan (j.duan@bham.ac.uk).}}
\maketitle

\begin{abstract}
Fibre orientation distribution (FOD) reconstruction using deep learning has the potential to produce accurate FODs from a reduced number of diffusion-weighted images (DWIs), decreasing total imaging time. Diffusion acquisition invariant representations of the DWI signals are typically used as input to these methods to ensure that they can be applied flexibly to data with different b-vectors and b-values; however, this means the network cannot condition its output directly on the DWI signal. In this work, we propose a spherical deconvolution network, a model-driven deep learning FOD reconstruction architecture, that ensures intermediate and output FODs produced by the network are consistent with the input DWI signals. Furthermore, we implement a fixel classification penalty within our loss function, encouraging the network to produce FODs that can subsequently be segmented into the correct number of fixels and improve downstream fixel-based analysis. Our results show that the model-based deep learning architecture achieves competitive performance compared to a state-of-the-art FOD super-resolution network, FOD-Net. Moreover, we show that the fixel classification penalty can be tuned to offer improved performance with respect to metrics that rely on accurately segmented of FODs. Our code is publicly available at \href{https://github.com/Jbartlett6/SDNet}{https://github.com/Jbartlett6/SDNet}.
\end{abstract}


\section{Introduction}
\label{sec:introduction}
\input{Sections/1.Introduction.tex}

\section{Method}
\label{sec:method}
\input{Sections/2.Method.tex}

\section{Experiments}
\label{sec:experiments}
\input{Sections/3.Experiments.tex}

\section{Results}
\label{sec:results}
\input{Sections/4.Results.tex}

\section{Discussion}
\label{sec:discussion}
\input{Sections/5.Discussion.tex}

\section{Conclusion}
\label{sec:conclusion}
\input{Sections/6.Conclusion}

\bibliographystyle{IEEEtranN}
\small\bibliography{bibliography}


\end{document}

%% file: Sections/1.Introduction.tex
\IEEEPARstart{F}{ibre} orientation distributions (FODs) relate signal attenuation in diffusion-weighted magnetic resonance images to the volume fractions and orientations of fibre populations in the brain \citep{tournier2004direct, tournier2007robust, jeurissen2014multi}. Their flexibility and capacity to discern intra-voxel fibre populations  facilitates a range of subsequent quantitative analyses; tractography algorithms can be used to obtain tractograms, and FOD segmentation can provide discrete fibre bundle elements (fixels) \citep{raffelt2012apparent, Raffelt2017}.

Multi-shell, high angular resolution diffusion imaging datasets are required to fit FODs with sufficient angular detail, and for separating the contribution of different tissue types \citep{Tournier2013, jeurissen2014multi}. The approximately linear relationship between the time a subject spends in the scanner and the number of diffusion-weighted images (DWIs) collected means acquiring such datasets is time consuming.

Deep learning can help to alleviate this issue by performing FOD reconstruction, the task of fitting high-fidelity FODs to a reduced number of DWI signals. To ensure their flexibility, such deep learning methods should be invariant to changes in diffusion MRI acquisition arising due to inter-facility variability or DWI volume corruption. Resampling techniques such as spherical harmonics (SH) \citep{koppers2016direct, elaldi2021equivariant, hosseini2022cttrack} and nearest neighbour \citep{karimi2021learning} interpolation have been explored to resample arbitrary DWI acquisitions onto a pre-defined spherical grid. Alternatively, an SH representation of the signal can be used as input to the network \citep{lin2019fast, nath2020deep, koppers2017reconstruction, jha2022vrfrnet}. FOD super-resolution methods \citep{patel2018better, lucena2021enhancing, zeng2022fod} perform constrained spherical deconvolution (CSD) as a pre-processing step and take the SH representation of the FOD as input. Results in the literature vary due to the range of acquisitions and CSD algorithms used to fit the FODs such as: single-shell-single-tissue \citep{patel2018better}, two-tissue \citep{lucena2021enhancing} and single-shell-three-tissue \citep{zeng2022fod} FODs.

High computational costs and the risk of overfitting mean it is not feasible to process all signals in the spatial and diffusion-acquisition dimensions concurrently. By predicting the central FOD from a limited spatial neighbourhood of the input \citep{lin2019fast, zeng2022fod, koppers2017reconstruction}, a compromise can be found between reducing the computational burden and exploiting the abundance of spatial correlations present in the data. Such methods commonly utilise a 3D convolutional neural network (CNN) for feature extraction, followed by fully connected or transformer layers for FOD prediction \citep{hosseini2022cttrack}. 

It is common practice for FODs to be fit using CSD with a maximum SH order of eight \citep{zeng2022fod, lin2019fast} in order to to capture angular frequency content of the DWI signal at a maximum b-value of $3000 \text{ s/mm}^{2}$ \citep{Tournier2013}. Some tractography algorithms require only the orientations of fibre populations in each voxel as input, so a number of FOD reconstruction algorithms predict only these quantities \citep{koppers2016direct, karimi2021learning}. Alternatively, an unsupervised loss function with sparsity inducing regularisation can be used to reconstruct FODs with an increased maximum order of 20 \citep{elaldi2021equivariant}. Whilst improving the angular separation, these methods change the FOD model, meaning it is likely that fixel-derived scalars, such as apparent fibre density and peak amplitude, also deviate. Therefore, it would be infeasible to apply such methods within a fixel-based analysis pipeline.

Model-based deep learning exploits domain knowledge of a process to inspire neural network architectures. Many approaches alternate between CNN-based denoising and data consistency blocks \citep{aggarwal2018modl, schlemper2017deep, jia2021learning, duan2019vs}. Data consistency blocks use prior knowledge of an appropriate forward model to ensure a network produces solutions consistent with the input signal. 

When calculating acquisition invariant representations of the DWI signal, fitting errors are incurred. We conjecture that such errors lead to the degradation of FOD reconstruction performance since the subsequently applied neural networks cannot directly condition their output on the true DWI signal, and model-based deep learning has the potential to lessen the impact of these errors by ensuring intermediate and output FODs are consistent with the DWI signal. In the context of FOD reconstruction, data consistency blocks minimise a linear combination of the CSD data consistency and an additional, deep learning based, regularisation term. Current implementations use a pre-trained autoencoder based regularisation term \citep{patel2018better}, however this means the network will not be optimised for FOD reconstruction performance. Model-based deep learning has to this point not been combined with techniques proven successful in end-to-end FOD reconstruction architecture.  

In this paper \textbf{S}pherical \textbf{D}econvolution \textbf{Net}work (SDNet) is introduced, a model-based deep learning architecture that utilises spatial information from surrounding voxels and is optimised to perform FOD reconstruction of multi-shell data. Additionally, we propose a fixel classification penalty within our loss function to improve angular separation without distorting the shape of the reconstructed FODs, which can be tuned to suit the requirements of the reconstructed FODs. The efficacy is evaluated by extensive comparisons with a state-of-the-art FOD super-resolution method, FOD-Net, as well as an ablation study. Our results show that including model-based deep learning improves the performance of the network.

%% file: Sections/2.Method.tex
\subsection{Network Architecture}
Constrained spherical deconvolution is used to fit FODs to DWI signal by optimising the following objective function:
\begin{equation}    
\label{eq:SDGen}
\mathop {\min\limits_{\bf{c}}} { {\frac{1}{2m}\| {{\cal A} {\cal Q}{\bf{c}} - {\bf{b}}} \|_2^2}  + {\cal R}\left( {\bf{c}} \right)}
\end{equation}

\begin{figure*}[!t]
\vspace{-10pt}
\centerline{\includegraphics[height=7cm, width=\textwidth]{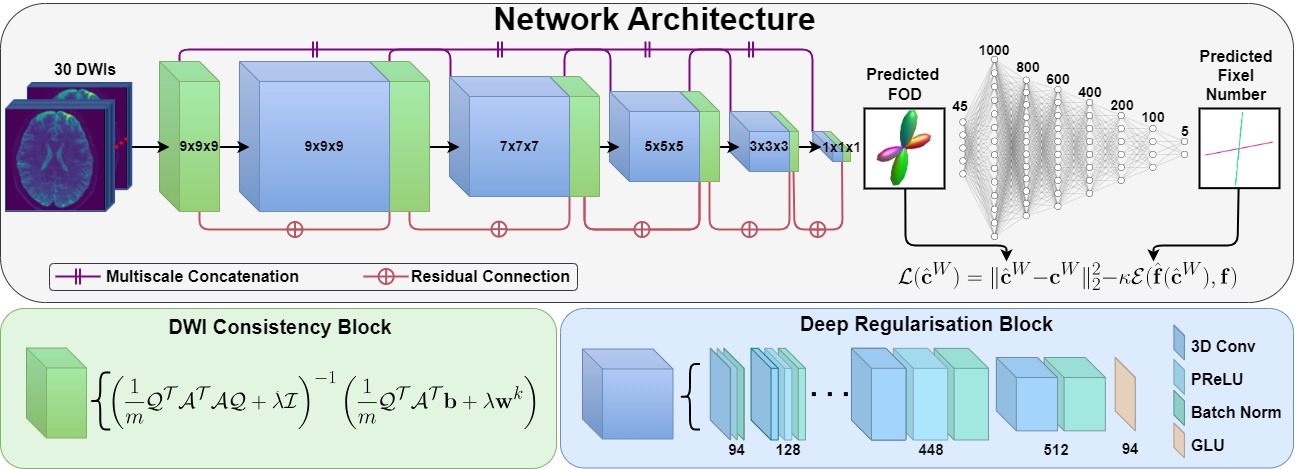}}
\caption{SDNet architecture, made up of alternating deep regularisation blocks and DWI consistency blocks. Each DWI consistency block is made up of 3D convolution blocks, the values above each set of layers represents the number of channels, which increase as follows: $\{94,128, 192, 256, 320, 384, 448\}$. The DWI consistency block shows the matrix inversion that is solved for each voxel independently.}
\label{fig:SDNet}
\vspace{-10pt}
\end{figure*}

\noindent where ${\bf{c}} \in \mathbb{R}^{n}$ are the SH coefficients of the FOD, ${\bf{b}}\in\mathbb{R}^{m}$ are the DWI signals, and ${\cal AQ} \in\mathbb{R}^{m \times n}$ spherically convolves the FOD with the response functions of the tissue types being modelled. To facilitate a data-driven regularisation term, optimised for FOD reconstruction, we consider an arbitrary regularisation term, ${\cal R}(\cdot)$, in place of the ubiquitous non-negativity constraint. In the following we outline how the variable splitting methods used in \citet{jia2021learning, duan2019vs} can be adapted to solve \eqref{eq:SDGen}.

First, we introduce an auxiliary splitting variable ${\bf{w}} \in \mathbb{R}^{n}$, converting \eqref{eq:SDGen} into the following equivalent form:
\begin{equation}
\label{eq:SDdecouplecon}
\mathop {\min\limits_{{\bf{c,w}}}} { {\frac{1}{2m}\| {{\cal A} {\cal Q}{\bf{c}} - {\bf{b}}} \|_2^2}  + {\cal R}\left( {\bf{w}} \right)} \; s.t. \; {\bf{c=w}}
\end{equation}

Using the penalty function method, we add these constraints back into the model and minimise the joint objective:
\begin{equation}
\label{eq:SDdecouple}
 \mathop { \min\limits_{\bf{c,w}}} { {\frac{1}{2m}\| {{\cal A} {\cal Q}{\bf{c}} - {\bf{b}}} \|_2^2}  + {\cal R}\left( {\bf{w}} \right)} + \frac{\lambda}{2}\|{\bf{c-w}}\|^{2}_{2}
\end{equation}

Eq.~(\ref{eq:SDdecouple}) can  be solved for ${\bf{c}}$ and ${\bf{w}}$ using an alternating optimisation scheme:
\begin{equation}
\label{SDalt}
\left\{ \begin{array}{l}{{\bf{c}}^{k + 1}} = \mathop {\arg \min\limits_{\bf{c}}} \frac{1}{2m} \| {{\cal A} {\cal Q}{\bf{c}} - {\bf{b}}} \|_2^2 + \frac{\lambda }{2} \| {\bf{c-w}}^{k} \|_2^2\\{\bf{w}}^{k + 1} = \mathop {\arg \min\limits_{{\bf{w}}}}  {\frac{\lambda}{2}\| {\bf{c}}^{k+1}-{\bf{w}} \|_2^2}  + {\cal R}\left( {\bf{c}}^{k+1} \right) \\\end{array} \right..
\end{equation}

The first convex optimisation can be solved using matrix inversion. The second equation is a denoising problem with arbitrary regularisation, the optimal form of which is unknown. In order to learn the regularisation to improve FOD reconstruction performance, the iterative process can be unrolled and the denoising step solved using a neural network, ${\cal{NN}}(\cdot)$:
\begin{equation}
\label{eq:SDunroll}
\left\{ \begin{array}{l}{{\bf{c}}^{k + 1}} = \left(\frac{1}{m} {\cal{Q}^{T}}{\cal{A}^{T}}{\cal{A}}{\cal{Q}} + \lambda {\cal{I}}\right)^{-1}\left(\frac{1}{m}{\cal{Q}^{T}}{\cal{A}^{T}}{\bf{b}} + \lambda {\bf{w}}^{k}\right)\\
{\bf{w}}^{k + 1} = {\cal{NN}}\left({\bf{c}}^{k+1}\right)\end{array} \right..
\end{equation}

The network architecture (Fig.~\ref{fig:SDNet}) takes nine voxels in each spatial dimension for 30 different diffusion gradients, resulting in a $9 \times 9 \times 9 \times 30$ volume of DWI signals as input, and passes them through alternating DWI consistency and deep regularisation blocks. The network outputs a vector ${\hat{\bf{c}}} \in \mathbb{R}^{n}$, a high-fidelity prediction of the FOD from the central voxel of the $9\times9\times9$ input patch. 

 \subsubsection{DWI Consistency}
Each DWI consistency block solves the matrix inversion in (\ref{eq:SDunroll}) independently for each voxel, maintaining spatial resolution. The initial DWI consistency block optimises only for the first three even orders of spherical harmonic coefficients $(l_{max}=4)$ to ensure robustness to aggressive DWI undersampling. 

\subsubsection{Deep Regularisation}
Each deep regularisation block is applied to a concatenation of the previous two DWI consistency blocks, meaning the block is conditioned on  both earlier representations. Validation tests showed these connections improve network performance (data not shown). The initial $3 \times 3 \times 3$ convolution kernels are applied with one layer of zero padding in each dimension, as to maintain spatial resolution, and are followed by 3D batch normalisation layers and parametric rectified linear unit (PReLU) activation functions. The number of channels is increased in this manner until it has reached 448 (Fig.~\ref{fig:SDNet}). No padding is applied in the final $3 \times 3 \times 3$ convolution kernel followed by a PReLU function, reducing the resolution in each spatial dimension by two. Finally, a $1 \times 1 \times 1$ convolution kernel is then applied to the 512-channel feature maps to obtain a 94-channel input to a gated linear unit (GLU) activation function,which is the output of the block. Residual connections, referencing the output of the previous DWI consistency block, are used to improve gradient flow through the network. The deep regularisation block reduces each spatial dimension of its input by two.

\subsection{Loss Functions}
In addition to the customary MSE loss, a fixel classification penalty is proposed to give greater control over the angular separation of the reconstructed FODs. The mechanics of this method can be considered similar to the microstructure sensitive loss proposed for DWI signal reconstruction in \citep{chen2023deep}.
To overcome the inherent non-differentiable nature of the fast marching level set FOD segmentation algorithm \citep{smith2013sift}, a fixel classification network is applied to predict the number of fixels each voxel contains. The output is passed into a cross-entropy component of the loss function. Since we are concerned with the white matter components of the FODs, the loss function and performance metrics are not functions of the grey matter and cerebrospinal fluid components of the FOD. For notational simplicity, from this point onwards ${\bf{c}}$ refers only to the white matter component of the FOD.  The overall loss function is as follows: 
\begin{equation}
{\cal{L}}(\hat{{\bf{c}}}) =\frac{1}{N_{batch}} \sum^{N_{batch}}_{i = 1} \left(\|\hat{{\bf{c}}}_{i} - {\bf{c}}_{i}\|^{2}_{2} + \kappa {\cal{E}}(\hat{{\bf{f}}}(\hat{{\bf{c}}}_{i}),{\bf{f}}_{i})\right)
\label{eq:Loss}
\end{equation}
where $N_{batch}$ is the number of data points in the mini-batch, $\hat{{\bf{c}}}_{i},\;{\bf{c}}_{i}\in\mathbb{R}^{n-2}$ are the reconstructed and fully sampled white matter FODs, ${\cal{E}}(\cdot,\cdot)$ is the cross-entropy, $\hat{{\bf{f}}}(\hat{{\bf{c}}}_{i}), \;{\bf{f}}_{i} \in \mathbb{R}^{5}$ are the predicted logits and the one-hot encoding of the number of fixels respectively and $\kappa$ is a hyperparameter to balance the two components of the loss function.

\begin{table}[h]
\caption{Count and percentage values of fixels in white matter voxels of an individual from the HCP dataset before and after thresholding at 4 fixels. Before thresholding there is a severe class imbalance.}
\label{tab:fixdist}
    \centering
    \begin{tabular}{?c?c?c?c?}
    \specialrule{1.25pt}{0pt}{0pt}
         \makecell{Number \\ of Fixels} & Count & \makecell{Percentage \\ before thresholding} & \makecell{Percentage \\ after thresholding} \\ \specialrule{1.25pt}{0pt}{0pt}
         1 & 310994 & 49\% & 49\% \\
         2 & 200673 & 32\% & 32\% \\
         3 & 76672 & 12\% & 12\% \\
         4 & 24095 & 4\% & 6.7\%\\
         5 & 10975 & 2\% & - \\
         6 & 4979 & 0.8\% & - \\
         7 & 1800 & 0.3\% & - \\
         \specialrule{1.25pt}{0pt}{0pt}
    \end{tabular}
\vspace{-0pt} 
\end{table}


 When training the fixel classification network, the number of fixels in each voxel were thresholded to four (Tab.~\ref{tab:fixdist}), reducing the inclusion of spurious peaks and class imbalance. A simple, fully-connected architecture was used, with layers containing $\{$45, 1000, 800, 600, 400, 200, 100, 5$\}$ neurons. Between each layer there are ReLU activation and 1D batch normalisation functions, other than between the penultimate and final layer where the batch normalisation is omitted. A softmax activation function, followed by cross-entropy loss, were then applied to the output of the network. The classification network was trained using the same training set as SDNet. Fully sampled FODs were used as the input, and the ground truth targets were calculated using the fast level set marching algorithm \citep{smith2013sift}. 
 
 \begin{figure*}[!htp]
 \vspace{-20pt}
\centerline{\includegraphics[ height=7cm, width=\textwidth]{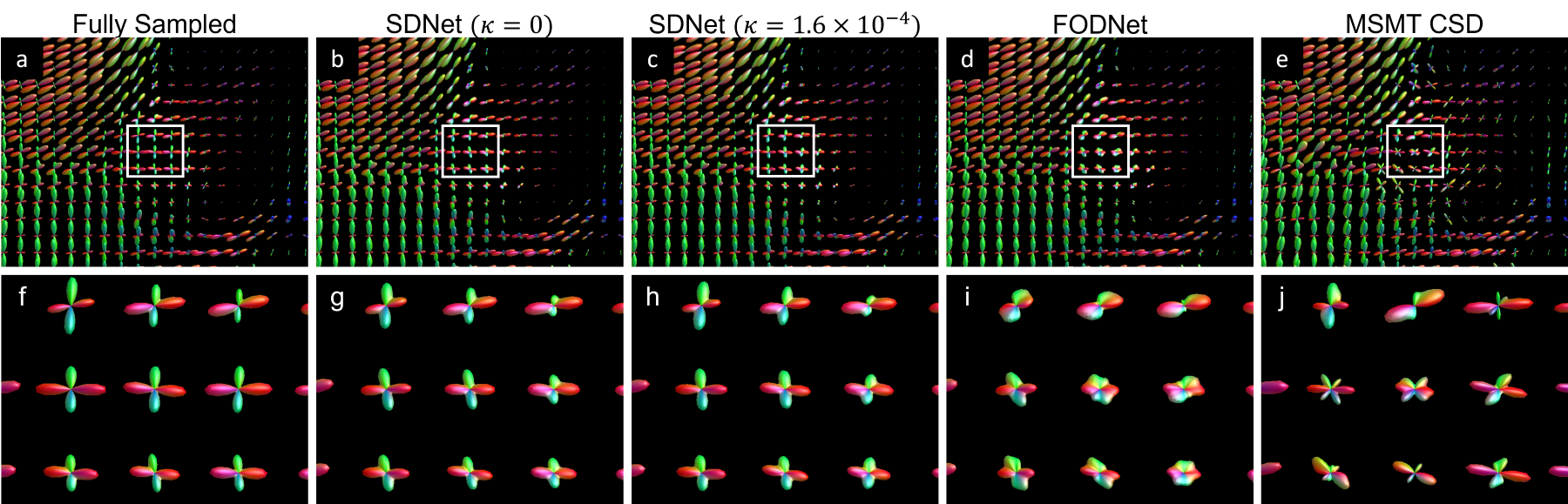}}
\caption{Qualitative results showing reconstructed FODs for HCP subject 130821 centred at voxel [38,98,70]. The top row consists of the \textbf{a.} Fully Sampled, \textbf{b.} SDNet, \textbf{c.} ${\text{SDNet}}_{\kappa}$, \textbf{d.} FOD-Net, and \textbf{e.} MSMT CSD FODs. The bottom row shows a zoomed-in area of FODs, corresponding to the region highlighted by the white square, consisting of the \textbf{f.} Fully Sampled, \textbf{g.} SDNet, \textbf{h.} ${\text{SDNet}}_{\kappa}$, \textbf{i.} FOD-Net, and \textbf{j.} MSMT CSD FODs. Where $\kappa$ is the hyperparameter which balances the SH error and fixel classification penalty terms in the loss function as per Eq.~\ref{eq:Loss}.
}
\label{fig:FODQual}
\vspace{-10pt}
\end{figure*}

\subsection{Implementation Details}
To demonstrate the impact of the fixel classification penalty, experiments were carried out with $\kappa = 0$ and $ \kappa = 1.6 \times 10^{-4}$. The ADAM optimiser \citep{kingma2014adam}, with learning rate warm-up, was used for parameter optimisation, with an initial learning rate of $10^{-6}$, increasing to $10^{-4}$ after $10,000$ iterations. To minimise hyperparameter tuning, $\lambda$ was optimised simultaneously with the network weights. From validation experiments (data not included), we found that the most effective way to utilise the classification loss to train SDNet was to initially train the model with $\kappa=0$ and then to increase $\kappa$ to its final value after this initial training stage. To do so we trained SDNet with only MSE loss until convergence, then trained the network until convergence with $ \kappa = 1.6 \times 10^{-4}$.

%% file: Sections/3.Experiments.tex
\subsection{Dataset}
A subset of the WU-Minn Human Connectome Project (HCP) dataset \citep{van2013wu}, consisting of 30 subjects, was split $20/3/7$ and used for training, validation, and testing, respectively. The HCP images have $1.25\text{mm}$ isotropic resolution with 90 gradient directions for $b = 1000, 2000 \text{ and } 3000 \text{ s/mm}^{2}$ and 18 $b_{0}$ images. The HCP dataset was minimally pre-processed in accordance with \citep{sotiropoulos2013advances}. 

Additionally, prior to applying SDNet, each subject's data was normalised using MRtrix3's \textit{dwinormalise} function. The fully sampled FODs were fit to all 288 DWIs; first, the response functions were calculated using the method proposed in \citep{dhollander2019improved}, then the FODs calculated using MSMT-CSD \citep{jeurissen2014multi}. White matter response functions and FODs were modelled with $l_{max} = 8$ and the grey matter and cerebrospinal fluid component response functions and FODs were modelled with $l_{max} = 0$, resulting in a total of 47 SH coefficients. 

The sampling pattern \citet{caruyer2013design} utilised in the HCP is such that for any $k$, selection of the first $k$ DWI volumes results in evenly spread b-vectors. To prepare the input data, the first 9 DWIs from each non-zero shell were selected with an additional 3 $b_{0}$ images, resulting in a total of 30 DWI signals. 

\begin{figure*}[t!]
\vspace{-15pt}
\centerline{\includegraphics[ width=0.7\textwidth]{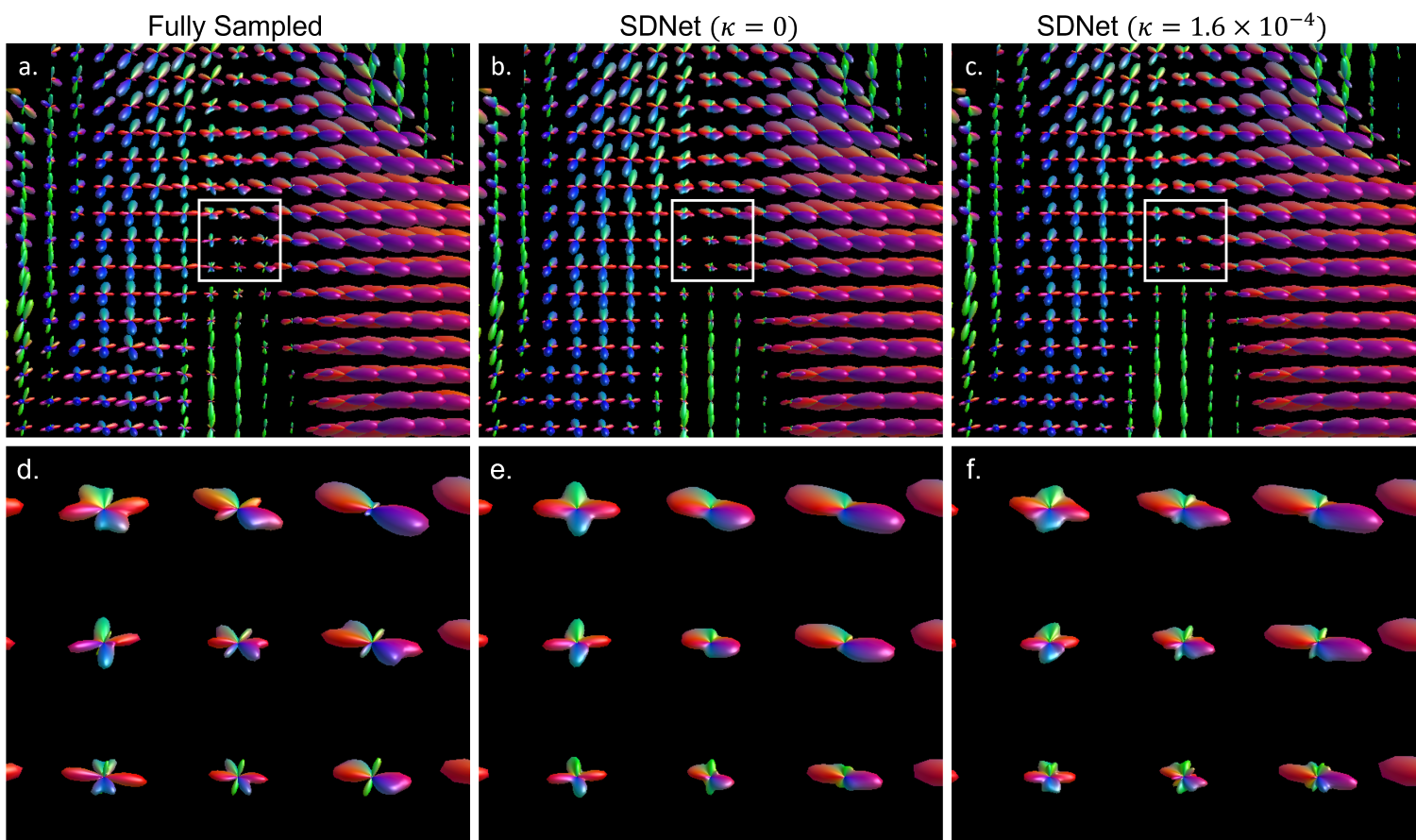}}
\caption{FODs taken from HCP subject 130821, centred at voxel [84,110,70]. The top row are the \textbf{a.} Fully Sampled, \textbf{b.} SDNet, and \textbf{c.} ${\text{SDNet}}_{\kappa}$ FODs. The second row consists a zoomed-in region of FODs, corresponding to the region highlighted by the white square the contain the \textbf{d.} Fully Sampled, \textbf{e.} SDNet, and \textbf{f.} ${\text{SDNet}}_{\kappa}$. 
FODs. Where $\kappa$ balances the SH error and fixel classification penalty terms in the loss function as per Eq.~\ref{eq:Loss}.
}
\label{fig:FixQual}
\vspace{-10pt}
\end{figure*}

Only patches in which the central voxel is classified as grey matter or white matter are used for training. The grey matter voxels were included to improve performance near the boundary of the two tissue types, as highlighted in \citep{zeng2022fod}. The grey and white matter masks were calculated using the method outlined in \citep{zhang2001segmentation}, which is implemented using the FSL software package \citep{jenkinson2012fsl}. 

From this point onwards, for notational convenience, SDNet ($\kappa = 0$) will be referred to as SDNet and SDNet ($\kappa = 1.6\times 10^{-4}$) will be referred to as ${\text{SDNet}}_{\kappa}$. To evaluate the performance of the introduced methods, SDNet, ${\text{SDNet}}_{\kappa}$, FOD-Net \citep{zeng2022fod}, and super-resolved MSMT CSD, referred to as MSMT CSD for notational simplicity, were all compared. In the original implementation, FOD-Net maps FODs fit using the single shell three tissue CSD algorithm \citep{dhollander2019improved} to 32 DWIs (4 $b_{0} \text{ and } 28 \text{ } b = 1000/2000/3000 \text{ s/mm}^{2}$) to the desired MSMT CSD obtained FODs. To allow a fair comparison between FOD-Net and the proposed networks, FOD-Net was trained using the same training set as SDNet. Since the final block in the SDNet architecture is a DWI consistency block, it cannot map to normalised FODs, therefore the target training data is not normalised. It should be noted that the normalisation can still be performed as a post-processing step. Otherwise, the same configuration settings found in the Github repository released by the FOD-Net authors were used.

\subsection{Performance Metrics}

To evaluate the performance of the FOD reconstruction algorithms, performance metrics were calculated voxel-wise then averaged over regions of interest.  The regions considered were the white matter and intersections of individual tracts within the white matter. The tracts considered were: the corpus callosum (CC), the middle cerebellar peduncle (MCP), the corticospinal tract (CST), and the superior longitudinal fascicle (SLF). To understand how the algorithm performs in voxels containing different numbers of fibres, we considered the intersections of these tracts as in \citep{zeng2022fod}. For voxels containing a single fibre, we considered voxels in the CC containing a single fixel, which we refer to as ROI-1-CC. For two crossing fibres, we considered voxels in the intersection of the MCP and CST containing two fixels, which we refer to as ROI-2-MCP. For three crossing fibres, we considered voxels in the intersection of the SLF, CST and CC containing three fixels, which we refer to as ROI-3-SLF. The white matter mask was calculated using the FSL five tissue type segmentation algorithm in MRtrix3. The segmentation masks for the white matter fibre tracts were obtained using TractSeg \citep{wasserthal2018tractseg}. 

The SSE between the reconstructed FODs, $\hat{{\bf{c}}}$, and the fully sampled FODs, ${\bf{c}}$, was computed as follows:  
\begin{equation}
\label{eq:SSE}
\text{SSE}\left({\bf{c}}, \hat{{\bf{c}}}\right) = \left\|{\bf{c}} - \hat{{\bf{c}}}\right\|^{2}_{2}
\end{equation}

The angular correlation coefficient (ACC) \citep{anderson2005measurement} was computed as follows:

\begin{equation}
\text{ACC}({\bf{c}}, \hat{{\bf{c}}}) = 
\frac{\sum\limits^{4}_{i=1}\sum\limits_{j=-2i}^{2i}{\bf{c}}_{2i,j}\hat{{\bf{c}}}_{2i,j}}{\sqrt{\left(\sum\limits^{4}_{i=1}\sum\limits^{2i}_{j=-2i}{\bf{c}}_{2i,j}^{2}\right)
\left(\sum\limits^{4}_{i=1}\sum\limits^{2i}_{j=-2i}\hat{{\bf{c}}}_{2i,j}^{2}\right)}}
\end{equation}

We refer to SSE and ACC as \textit{FOD-based performance} metrics, since they compare the SH representation of the FODs prior to any further processing. 

Fixel-based analysis requires each FOD to be segmented into fixels, each of which has associated apparent fibre density and peak amplitude \citep{smith2013sift}. To calculate the associated error metrics, peak amplitude and apparent fibre density vectors must be assembled. Each vector consists of the respective scalar for each fixel ordered according to the peak amplitude and are padded to a fixed length. The remaining metrics are referred to as \textit{fixel-based performance} metrics since they require the FOD to be segmented into fixels prior to evaluation. 

Fixel accuracy was defined for a region of interest as the proportion of voxels in which the FOD is segmented into the correct number of fixels.  

The peak amplitude error (PAE) was calculated between the reconstructed, $\hat{{\boldsymbol{f}}}^{P}$, and fully sampled FOD's, ${\boldsymbol{f}}^{P}$, peak amplitude vectors:
\begin{equation}
\label{eq:PAE}
\text{PAE}\left({\boldsymbol{f}}^{P}, \hat{{\boldsymbol{f}}}^{P}\right) = \sum\limits_{i}\left|f^{P}_{i}-\hat{f}^{P}_{i}\right|
\end{equation}

The apparent fibre density error (AFDE) was calculated between the reconstructed, $\hat{{\boldsymbol{f}}}^{A}$, and fully sampled FOD's, ${\boldsymbol{f}}^{A}$, apparent fibre density vectors:
\begin{equation}
\label{eq:AFDE}
\text{AFDE}\left({\boldsymbol{f}}^{A}, \hat{{\boldsymbol{f}}}^{A}\right) = \sum\limits_{i}\left|f^{A}_{i}-\hat{f}^{A}_{i}\right|
\end{equation}




\subsection{Ablation Study}
To investigate the impact of the DWI consistency block on the performance of the network, an ablation study was conducted. The network was trained without the DWI consistency blocks, and all other aspects of the architecture and network training remained the same. We compared this model to SDNet with the DWI consistency blocks included.


\begin{figure*}[t!]
\vspace{-15pt}
\centerline{\includegraphics[ width=0.83\textwidth]{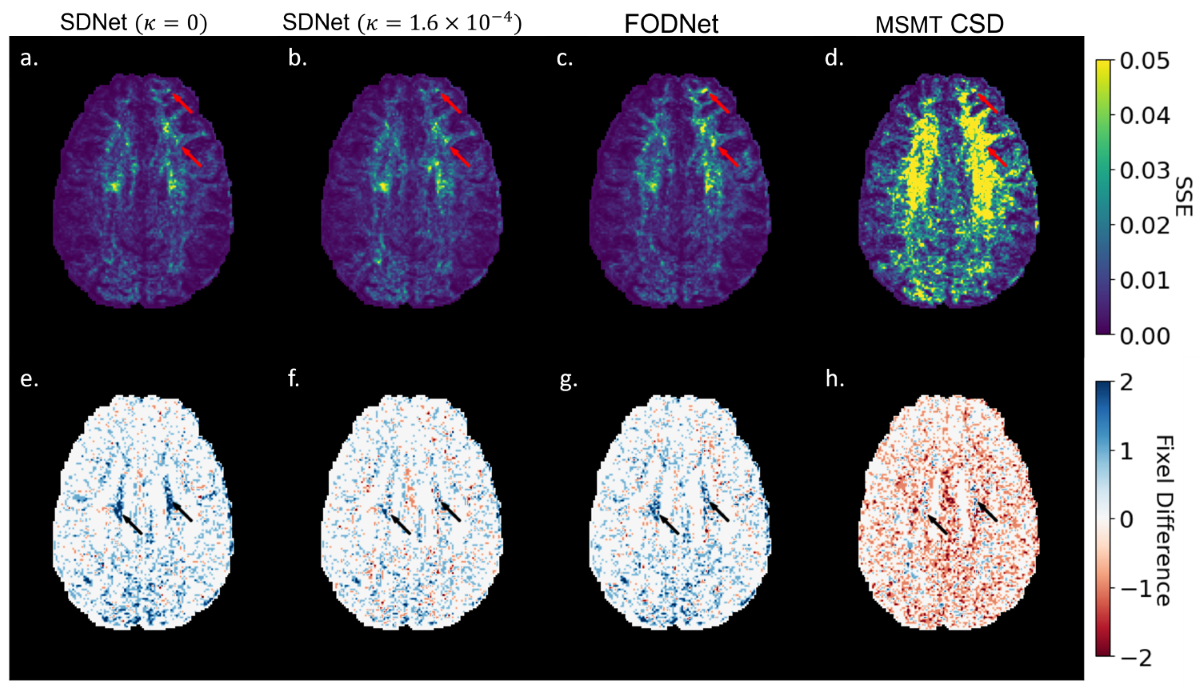}}
\caption{SSE and fixel difference error maps for slice 72 from HCP subject 130821. \textbf{Top row:} SSE error maps between the fully sampled FODs and the FODs reconstructed by \textbf{a.} SDNet, \textbf{b.} ${\text{SDNet}}_{\kappa}$, \textbf{c.} FOD-Net and \textbf{d.} MSMT CSD. \textbf{Bottom row:} Number of fixels calculated for the fully sampled FOD minus the number of fixels calculated from the FODs reconstructed by \textbf{e.} SDNet, \textbf{f.} ${\text{SDNet}}_{\kappa}$, \textbf{g.} FOD-Net, and \textbf{h.} MSMT CSD. Where $\kappa$ balances the SH error and fixel classification penalty terms in the loss function as per Eq.~\ref{eq:Loss}. Blue voxels indicate underestimates, and red areas overestimates, of the number of fixels. Large SSE and fixel differences are highlighted by the black and red arrows respectively.}
\label{fig:ERRmaps}
\vspace{-15pt}
\end{figure*}

\subsection{Statistical Analysis}
Shapiro-Wilk tests for normality ($\alpha=0.05$) were applied for each performance metric and method; unless otherwise stated there is insufficient evidence to reject the null hypothesis that the groups are normally distributed.

Since the data was normally distributed, and each method was applied to the same set of test subjects, a repeated measures one-way ANOVA ($\alpha=0.05$) was applied to each performance metric to determine whether there was a main effect between the conditions. 
Finally, to determine which methods contributed to the main effect, post-hoc t-tests with Bonferroni correction (adjusted for $\alpha = 0.05$) were used to identify effects between the FOD reconstruction algorithms.

%% file: Sections/4.Results.tex


\subsection{Qualitative Results}
The qualitative results comparing all methods (Fig.~\ref{fig:FODQual}) show that the deep learning methods reconstructed FODs that more closely resembled the ground truth when compared to MSMT CSD. The primary difference is the presence of spurious peaks produced by MSMT CSD, whereas the deep learning based algorithms coherently captured the major tracts in this region due to their denoising effect. 

The highlighted region in Fig.~\ref{fig:FODQual} (panels \textbf{f.}-\textbf{j.}) shows an area where FOD-Net produced distorted FODs compared to SDNet and ${\text{SDNet}}_{\kappa}$. MSMT CSD reconstructed particularly noisy FODs in this area, which the results obtained by FOD-Net resembled some similarities to. The FODs produced by SDNet underestimated the amplitude in this region but more accurately distinguished between fibre populations and captured their directions. In this region, which contains dominant fibre populations with large angular separation, the impact of increasing $\kappa$ on the reconstructed FODs is minimal; only a small change in the direction of the fibres is observed. In the larger tracts in panels Fig.~\ref{fig:FODQual} \textbf{a.}-\textbf{e.}, such as the green fibre population going upwards in the bottom left corner, all deep learning methods performed similarly. 

The qualitative results comparing SDNet with ${\text{SDNet}}_{\kappa}$ (Fig.~\ref{fig:FixQual}) illustrate that ${\text{SDNet}}_{\kappa}$ better separated fibre populations. The presence of fibre populations going from the lower left to upper right of panels Fig.~\ref{fig:FixQual} \textbf{d.}-\textbf{f.} are separated from the larger fibre population by ${\text{SDNet}}_{\kappa}$ but not by SDNet without the fixel classification penalty. The FODs reconstructed in the broader region, captured in panels Fig.~\ref{fig:FixQual} \textbf{a.}-\textbf{c.}, show that  larger fibre populations are reconstructed similarly for both SDNet and  ${\text{SDNet}}_{\kappa}$.

\subsection{FOD-based Results}
\begin{figure*}[htp]
\vspace{-5pt}
\centerline{\includegraphics[ width=0.88\textwidth]{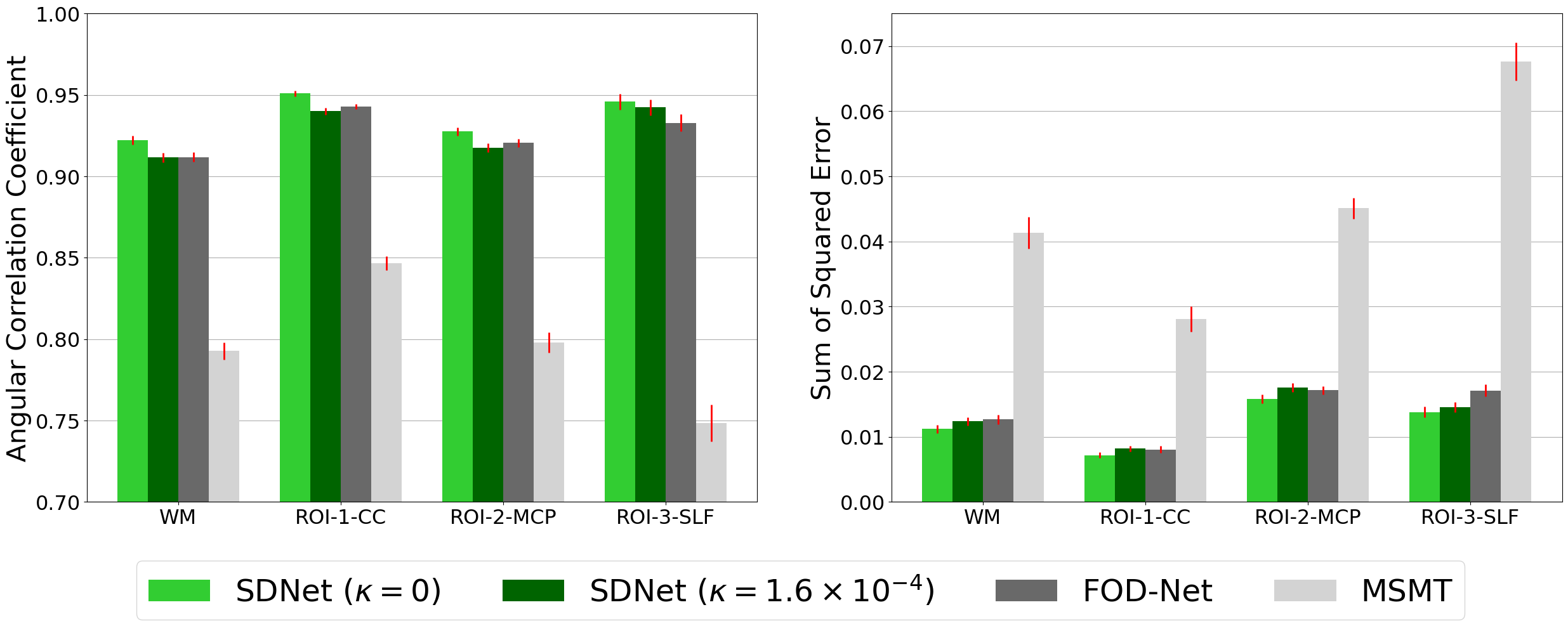}}
\caption{Mean test-time FOD-based performance (\textbf{Left:} ACC, \textbf{Right:} SSE) in the white matter (WM), ROI-1-CC: corpus callosum containing a single fixel, ROI-2-MCP: intersection between the middle cerebellar peduncle and superior longitudinal fascicle containing 2 fixels, and ROI-3-SLF: intersection between the superior longitudinal fascicle, corticospinal tract and the corpus callosum containing 3 fixels. $\kappa$ balances the SH error and fixel classification penalty terms in the loss function as per Eq.~\ref{eq:Loss}. Error bars indicate the standard error of the metrics, which have been averaged over the 7 test subjects.}
\label{fig:FODperf}
\vspace{-5pt} 
\end{figure*}


The SSE error maps (Fig.~\ref{fig:ERRmaps}) show that lower SSE is achieved throughout the brain by all deep learning methods compared to MSMT CSD. SDNet generally achieved smaller errors than the other deep learning methods. This is particularly evident in, but not restricted to, the areas highlighted by the red arrows. The error maps produced by ${\text{SDNet}}_{\kappa}$ and FOD-Net are similar.

 The average FOD-based performance results (Fig.~\ref{fig:FODperf} and Tab.~\ref{tab:QuantRes}) show that SDNet reconstructed FODs with significantly lower SSE and higher ACC than the compared methods in all regions of interest considered. The training curves (Fig.~\ref{fig:TrainCurve}) show that increasing $\kappa$ caused the validation ACC to decrease over the validation set. 

 In the white matter voxels, SDNet achieved the lowest SSE by a statistically significant margin over all compared methods, followed by ${\text{SDNet}}_{\kappa}$ and FOD-Net, between which there was no statistically significant difference in SSE. SDNet also achieved the strongest ACC performance in the white matter, where it improved over all other methods by a statistically significant margin. There was no statistically significant difference between ${\text{SDNet}}_{\kappa}$ and FOD-Net with respect to ACC in the white matter. 

In all of ROI-1-CC, ROI-2-MCP, and ROI-3-SLF, SDNet achieved the strongest SSE and ACC results (Fig.~\ref{fig:FODperf} and Tab.~\ref{tab:QuantRes}) by a statistically significant margin. FOD-Net and ${\text{SDNet}}_{\kappa}$ showed no statistically significant differences with respect to SSE and ACC in ROI-1-CC and ROI-2-MCP but in ROI-3-SLF ${\text{SDNet}}_{\kappa}$ achieved a statistically significant improvement over FOD-Net with respect to both SSE and ACC. In all regions, all deep learning based FOD reconstruction methods outperformed MSMT CSD with respect to SSE and ACC by a statistically significant margin. 

 

\subsection{Fixel-based Results}
\begin{figure}[h]
\centerline{\includegraphics[ width=0.9\columnwidth]{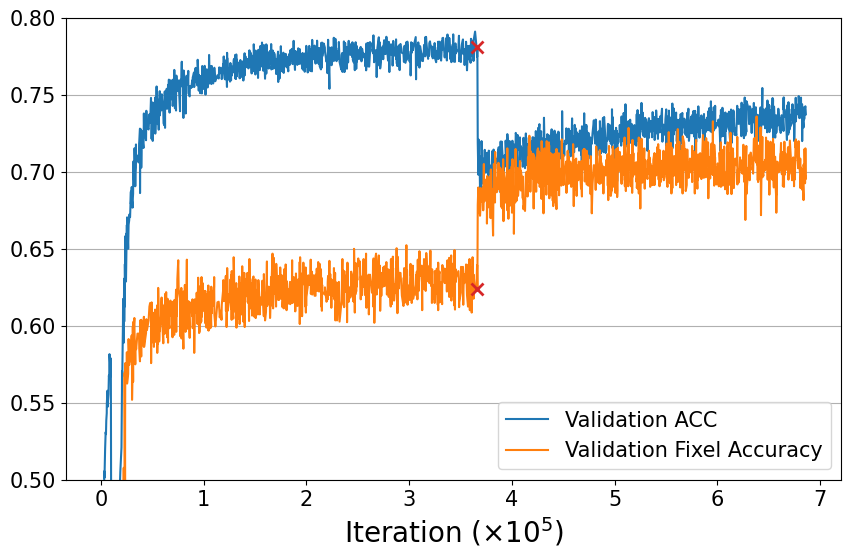}}
\vspace{-5pt}
\caption{Validation training curves for SDNet, the red cross marks the point when $\kappa$ is increased from $0$ to $1.6\times10^{-4}$}
\label{fig:TrainCurve}
\vspace{-15pt}
\end{figure}

 The fixel-based performance results (Fig.~\ref{fig:Fixperf} and Tab.~\ref{tab:QuantRes}) show greater variation between regions and an increased dependence on $\kappa$. The training curves (Fig.~\ref{fig:TrainCurve}) show that increasing $\kappa$ caused the validation fixel accuracy to increase over the validation set. In the white matter, ${\text{SDNet}}_{\kappa}$ achieved the strongest fixel accuracy by a significant margin, followed by SDNet and FOD-Net between which there was no statistically significant difference.



In ROI-1-CC, ROI-2-MCP, and ROI-3-SLF, we see that the fixel accuracy of the deep learning FOD reconstruction methods decreased as the number of fixels increased. In ROI-1-CC, SDNet achieved the strongest performance by a statistically significant margin, followed by FOD-Net and ${\text{SDNet}}_{\kappa}$, between which there is no statistically significant difference in fixel accuracy in the same region.

As the number of fixels in the ROIs increased, the fixel accuracy of ${\text{SDNet}}_{\kappa}$ increased relative to other methods. In ROI-2-MCP, ${\text{SDNet}}_{\kappa}$ achieved the highest fixel accuracy but not by a statistically significant margin over FOD-Net. Both methods outperformed SDNet by a statistically significant margin.  In ROI-3-SLF this pattern continued as ${\text{SDNet}}_{\kappa}$'s performance further improved, and it achieved a statistically significant fixel accuracy increase over the other deep learning methods. There was no statistically significant difference in fixel accuracy between FOD-Net and SDNet in ROI-3-SLF. In all regions other than ROI-3-SLF, MSMT performed worse than all other methods by a statistically significant margin. 

For AFDE in the white matter, ${\text{SDNet}}_{\kappa}$ achieved the lowest error by a statistically significant margin, followed by FOD-Net and SDNet between which there is no statistically significant difference in AFDE in the white matter. For PAE in the white matter, ${\text{SDNet}}_{\kappa}$ achieved the lowest error, which was a statistically significant improvement over SDNet but not FOD-Net. For both AFDE and PAE in the white matter, MSMT CSD achieved a higher error than all compared methods by a statistically significant margin.

\input{Sections/res_tab}
 
In ROI-1-CC, ROI-2-MCP and ROI-3-SLF, both AFDE and PAE generally increased as the number of fixels increased. In ROI-1-CC, SDNet achieved strongest results with respect to both AFDE and PAE and in ROI-2-MCP all three deep learning methods performed similarly with respect to both AFDE and PAE. In ROI-3-SLF, SDNet and ${\text{SDNet}}_{\kappa}$ achieved similar AFDE and PAE, with no statistically significant difference between them, but both achieved a statistically significant improvement compared to FOD-Net.

\subsection{Ablation Study}
\begin{table} 
    \caption{The results of all five performance metrics (mean $\pm$ standard error), averaged over all white matter voxels in all 7 test subjects. \textbf{${\bf 2^{nd}}$ column:}  SDNet, \textbf{${\bf3^{rd}}$ column:} SDNet without the DWI consistency block, \textbf{${\bf 4^{th}}$ column:} percentage difference between SDNet with and without the DWI consistency blocks, \textbf{${\bf 5^{th}}$ column:} pairwise t-test p-values. Bold p-values indicate a significant ($\alpha = 0.05$) effect.}
\label{tab:ablation}
\resizebox{1\columnwidth}{!}{
    \begin{tabular}{?c?c|c|c|c?}
         \specialrule{1.25pt}{0pt}{0pt}
         Metric & SDNet & \makecell{SDNet \\ w/o DC} & \makecell{Percentage \\ Change} & \makecell{p\\value}\\
         \specialrule{1.25pt}{0pt}{0pt}
         SSE ($\downarrow$)& ${\bf{0.011 \pm 0.001}}$  & $0.012 \pm 0.001$ & 9.1 \%  & {$<$ \textbf{0.05}} \\
         \hline
         ACC ($\uparrow$) & ${\bf{92.209 \pm 0.003}}$ & $91.679 \pm 0.003$ & 0.57\% & {$<$ \textbf{0.05}}\\
         \hline
         Fix Acc ($\uparrow$)& ${\bf{0.640 \pm 0.011}}$ & $0.625 \pm 0.010$ & 2.3\% & {$<$ \textbf{0.05}}\\
         \hline
         AFDE ($\downarrow$) & ${\bf{0.164 \pm 0.005}}$ & $0.177 \pm 0.005$ & 1.3 \% & {$<$ \textbf{0.05}}\\
         \hline
         PAE ($\downarrow$)& ${\bf{0.155 \pm 0.006}}$ & $0.163 \pm 0.006$ & 0.8\% & {$<$ \textbf{0.05}}\\
         \specialrule{1.25pt}{0pt}{0pt}
    \end{tabular}
    }
    \vspace{-10pt}
\end{table}

The results of the ablation study (Tab.~\ref{tab:ablation}) clearly demonstrate that removing the DWI consistency blocks from the SDNet architecture caused the performance of the network to degrade significantly with respect to all metrics. The greatest relative degradation of performance occurred with respect to SSE, however consistent reductions in the performance of all other metrics was also observed. 

%% file: Sections/res_tab.tex
\begin{table*}
\caption{Quantitative results for all performance metrics, methods and regions of interest. The arrows below each metric represent the direction of improved performance. The p-values for the paired t-test results between the respective methods and SDNet and $\text{SDNet}_{\kappa}$ are indicated by $\text{p}_{SD}$ and $\text{p}_{SD \kappa}$, respectively. The performance metric of the strongest method in each row is bold, significant p-values (Bonferroni corrected, adjusted for $\alpha=0.05$) are also bold.}
\label{tab:QuantRes}
  \centering
  \resizebox{0.9\textwidth}{!}{%
  \begin{tabular}{?c?l?llll?}
    \specialrule{1.25pt}{0pt}{0pt}
    \textbf{Metric} &\diagbox[innerwidth=1.5cm]{\textbf{Region}}{\textbf{Method}}& 
    \textbf{SDNet}&
    \textbf{SDNet\textsubscript{\boldmath$\kappa$} (p\textsubscript{SD})}&
    \textbf{FOD-Net (p\textsubscript{SD}, p\textsubscript{SD\boldmath${\kappa}$})} & \textbf{MSMT CSD (p\textsubscript{SD}, {p}\textsubscript{SD\boldmath$\kappa$})}\\
    \specialrule{1.1pt}{0pt}{0pt}
    \multirow{4}{*}{\textbf{\shortstack{SSE\\($\downarrow$)}}} & 
    White Matter &
    \textbf{0.011$\pm$0.001} &
    0.012$\pm$0.001 ($<$\textbf{0.001})&
    0.013$\pm$0.001 ($<$\textbf{0.001}, 0.034)& 
    0.041$\pm$0.002 ($<$\textbf{0.001}, $<$\textbf{0.001}) \\
    &
    ROI-1-CC&
    \textbf{0.007$\pm$0.001}&
    0.008$\pm$0.001 ($<$\textbf{0.001}) &
    0.008$\pm$0.001 ($<$\textbf{0.001}, 0.295) &
    0.028$\pm$0.002 ($<$\textbf{0.001}, $<$\textbf{0.001}) \\
    &
    ROI-2-MCP &
     \textbf{0.016$\pm$0.001}&
    0.018$\pm$0.001 ($<$\textbf{0.001}) &
    0.017$\pm$0.001 (\textbf{0.001}, 0.175)&
    0.045$\pm$0.002 ($<$\textbf{0.001}, $<$\textbf{0.001})\\
    
    &
    ROI-3-SLF & 
    \textbf{0.014$\pm$0.001}&
    0.015$\pm$0.001 (\textbf{0.005})& 
    0.017$\pm$0.001 ($<$\textbf{0.001}, $<$\textbf{0.001})& 
    0.063$\pm$0.002 ($<$\textbf{0.001}, $<$\textbf{0.003})\\
    
    \specialrule{1.25pt}{0pt}{0pt}
    \multirow{4}{*}{\textbf{\shortstack{ACC\\($\uparrow$)}}} &
    White Matter &
    \textbf{92.209$\pm$0.003}&
    91.152$\pm$0.003 ($<$\textbf{0.001}) &
    91.184$\pm$0.003 ($<$\textbf{0.001}, 0.484)&
    79.268$\pm$0.005 ($<$\textbf{0.001}, $<$\textbf{0.001})\\
    
    &
    ROI-1-CC&
    \textbf{95.090$\pm$0.002}&
    93.994$\pm$0.002 ($<$\textbf{0.001})&
    94.297$\pm$0.002 ($<$\textbf{0.001}, 0.009)&
    84.662$\pm$0.004 ($<$\textbf{0.001}, $<$\textbf{0.001})\\
    
    &
    ROI-2-MCP &
    \textbf{92.762$\pm$0.003}&
    91.746$\pm$0.003 ($<$\textbf{0.001})& 
    92.046$\pm$0.003 ($<$\textbf{0.001}, 0.032)& 
    79.796$\pm$0.006 ($<$\textbf{0.001}, $<$\textbf{0.001})\\
    
    &
    ROI-3-SLF &
    \textbf{94.577$\pm$0.005}& 
    94.233$\pm$0.005 (\textbf{0.001})&
    93.291$\pm$0.005 ($<$\textbf{0.001}, $<$\textbf{0.001})&
    74.844$\pm$0.011 ($<$\textbf{0.001}, $<$\textbf{0.001})\\
    \specialrule{1.25pt}{0pt}{0pt}
    \multirow{4}{*}{\shortstack{\textbf{Fix}\\
    \textbf{Acc}\\
    ($\uparrow$)}} &
    White Matter &
    0.640$\pm$0.011& 
    \textbf{0.664$\pm$0.008} ($<$\textbf{0.001})&
    0.645$\pm$0.009 (0.037, $<$\textbf{0.001})&
    0.536$\pm$0.006 ($<$\textbf{0.001}, $<$\textbf{0.001})\\
    
    &
    ROI-1-CC&
    \textbf{0.901$\pm$0.002}& 
    0.851$\pm$0.005 ($<$\textbf{0.001})&
    0.867$\pm$0.003 ($<$\textbf{0.001}, 0.036)&
    0.469$\pm$0.010 ($<$\textbf{0.001}, $<$\textbf{0.001})\\
    
    &
    ROI-2-MCP &
    0.754$\pm$0.011& 
    \textbf{0.791$\pm$0.010} ($<$\textbf{0.001})&
    0.772$\pm$0.009 (\textbf{0.001}, 0.018)&
    0.548$\pm$0.009 ($<$\textbf{0.001}, $<$\textbf{0.001})\\
    
    &
    ROI-3-SLF &
    0.606$\pm$0.032& 
    \textbf{0.648$\pm$0.031} (\textbf{0.001})&
    0.588$\pm$0.029 (0.023, $<$\textbf{0.001})&
    0.548$\pm$0.009 (0.076, 0.163)\\
    \specialrule{1.25pt}{0pt}{0pt}
    \multirow{4}{*}{\textbf{\shortstack{PAE \\  ($\downarrow$)}}} &
    White Matter &
    0.155$\pm$0.006& 
    \textbf{0.147$\pm$0.005} (\textbf{0.001})&
    0.152$\pm$0.005 (0.065, 0.011)&
    0.244$\pm$0.007 ($<$\textbf{0.001}, $<$\textbf{0.001})\\
    
    &
    ROI-1-CC&
    \textbf{0.062$\pm$0.002}& 
    0.072$\pm$0.002 ($<$\textbf{0.001})&
    0.069$\pm$0.002 ($<$\textbf{0.001}, 0.053)&
    0.210$\pm$0.007 ($<$\textbf{0.001}, $<$\textbf{0.001})\\
    
    &
    ROI-2-MCP &
     \textbf{0.135$\pm$0.003}& 
     0.136$\pm$0.003 (0.393)&
     0.136$\pm$0.002 (0.843, 0.973)&
     0.219$\pm$0.004 ($<$\textbf{0.001}, $<$\textbf{0.001})\\
    &
    ROI-3-SLF &
     0.179$\pm$0.009& 
     \textbf{0.178$\pm$0.007} (0.779)&
     0.194$\pm$0.010 (\textbf{0.001}, \textbf{0.002})&
     0.278$\pm$0.006 ($<$\textbf{0.001}, $<$\textbf{0.001})\\
    \specialrule{1.25pt}{0pt}{0pt}
    \multirow{4}{*}{\textbf{\shortstack{AFDE \\ ($\downarrow$)}}} &
    White Matter&
    0.164$\pm$0.005& 
    \textbf{0.151$\pm$0.004} ($<$\textbf{0.001})&
    0.160$\pm$0.005 (0.012, \textbf{0.002})&
    0.208$\pm$0.006 ($<$\textbf{0.001}, $<$\textbf{0.001})\\
    
    &
    ROI-1-CC&
    \textbf{0.065$\pm$0.001}& 
    0.074$\pm$0.001 ($<$\textbf{0.001})&
    0.073$\pm$0.002 (\textbf{0.002}, 0.711)&
    0.187$\pm$0.007 ($<$\textbf{0.001}, $<$\textbf{0.001})\\
    
    &
    ROI-2-MCP &
     0.107$\pm$0.002& 
     \textbf{0.105$\pm$0.001} (0.526)&
     0.106$\pm$0.001 (0.713, 0.489)&
     0.171$\pm$0.003 ($<$\textbf{0.001}, $<$\textbf{0.001})\\
    
    &
    ROI-3-SLF&
     0.151$\pm$0.007& 
     \textbf{0.149$\pm$0.006} (0.462)&
     0.165$\pm$0.007 ($<$\textbf{0.001}, $<$\textbf{0.001})&
     0.230$\pm$0.006 ($<$\textbf{0.001}, $<$\textbf{0.001})\\
    \specialrule{1.25pt}{0pt}{0pt}
\end{tabular}%

}
\end{table*}

%% file: Sections/5.Discussion.tex
SDNet is a model-based deep learning architecture that employs DWI consistency blocks to ensure intermediate FODs are consistent with the DWI signal, whilst making use of spatial information and multi-shell DWI data to reconstruct FODs. We compared our network to FOD-Net \citep{zeng2022fod}, a FOD super-resolution network, which fits FODs to the DWI signal prior to the network's forward pass. Our results show that SDNet improved over FOD-Net in terms of FOD-based performance, and performed similarly with respect to most fixel-based metrics. We conjecture that FOD-Net loses some details of the DWI signal in the FOD fitting stage. Our qualitative results (Fig.~\ref{fig:FODQual}) support this since the FODs reconstructed by FOD-Net more closely resembled the unstable input MSMT-CSD FODs, whereas by ensuring consistency with the DWI signal, SDNet more robustly reconstructed FODs which closely resembled the ground truth. The quantitative results collected from our comparison and ablation studies highlighted the improvement in FOD-based performance enabled by including DWI consistency blocks.

The ultimate goal of deep learning based FOD reconstruction is to produce FODs that are useful for quantitative analysis. FOD registration \citep{raffelt2011symmetric}, a key component of longitudinal and group FOD analyses, relies on $L_{2}$ distance between SH coefficients to captures FOD similarity. By achieving a low SSE, the SH representations will bear increased similarity to the ground truth FODs. We therefore anticipate that SDNet will help ensure that FOD registration is minimally impacted by DWI undersampling, and so too the subsequent analysis. 

\begin{figure*}[t!]
\vspace{-10pt}
\centerline{\includegraphics[ width=0.94\textwidth]{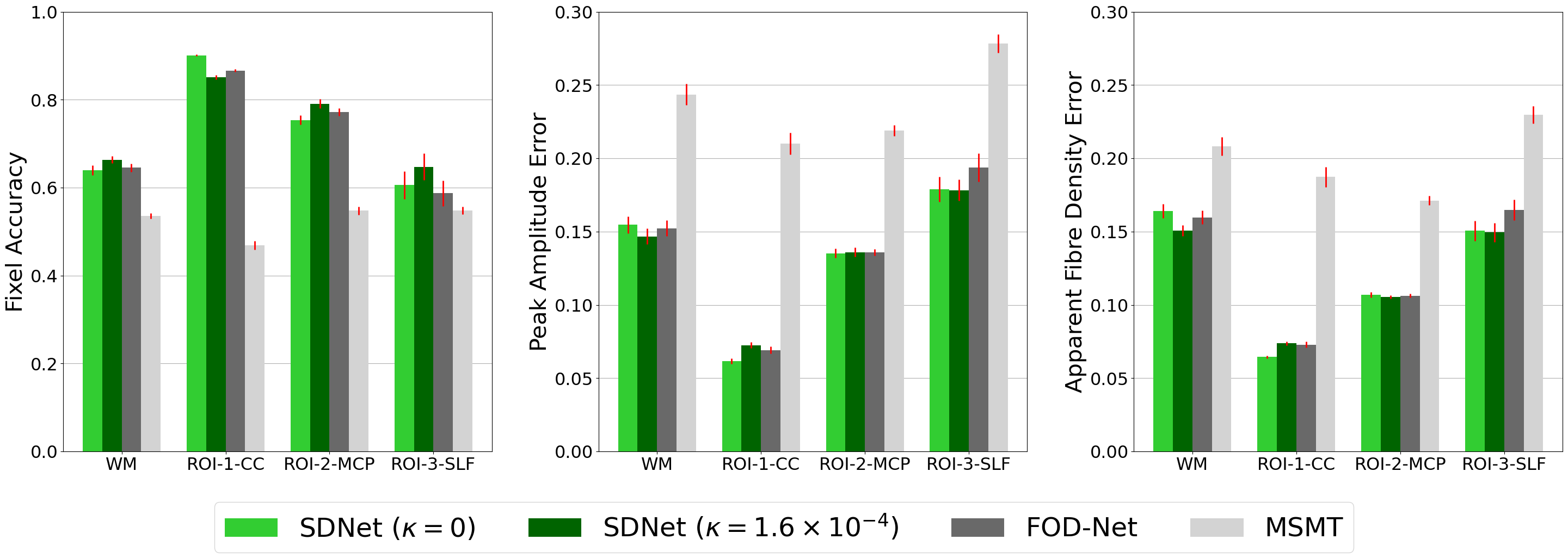}}
\caption{Mean test-time Fixel-based performance (\textbf{Left:} Fixel Accuracy, \textbf{Centre:} Apparent Fibre Density Error, \textbf{Right:} Peak Amplitude Error) in the white matter (WM), ROI-1-CC, voxels in the corpus callosum containing single fixels, ROI-2-MCP voxels in the intersection between the middle cerebellar peduncle and superior longitudinal fascicle containing 2 fixels, and ROI-3-SLF voxels in the intersection between the superior longitudinal fascicle, corticospinal tract and the corpus callosum containing 3 fixels. Error bars indicate the standard error of the metrics which have been averaged over the 7 test subjects.}
\label{fig:Fixperf}
\vspace{-10pt}
\end{figure*}

Another factor that may impact such analyses is data containing abnormalities, such as pathologies. Such data will likely not be abundant in the datasets used for training deep learning based FOD reconstruction networks, and as a consequence, reduced performance caused by overfitting becomes probable. Since the DWI consistency blocks ensure that solutions will be consistent with the measured DWI data, we expect that SDNet will be less likely to overfit therefore performing comparatively well compared to networks without DWI consistency blocks. However, further investigation is beyond the scope of the current work.

The outcome of such quantitative analysis is also dependent on the post-registration steps in the pipeline, which, in the case of a fixel-based analysis \citep{raffelt2012apparent}, will be predominantly impacted by the fixel-based performance. Comparing multiple FOD reconstruction algorithms revealed that strong FOD-based performance doesn't directly translate to strong fixel-based performance. The disconnect between FOD and fixel-based performance is evident in the statistically significant difference in SSE over the white matter between SDNet and FOD-Net, but the absence of a statistically significant effect in fixel accuracy over the same set of voxels. This effect can be attributed to FOD segmentation's dependence on the angular separation of the FOD lobes, which is dependent on the higher order SH coefficients, which only contribute a small amount to the SSE. This highlights that SSE loss alone may not be optimal for reconstructing FODs that are to be used in a fixel-based analysis pipeline. 


By introducing an additional loss component, which penalises reconstructed FODs judged to be made up of the incorrect number of fixels, we have demonstrated that fixel-based performance can be improved. The impact of the proposed loss function is illustrated by the statistically significant increase in fixel accuracy in the white matter achieved by $\text{SDNet}_{\kappa}$ compared to SDNet and FOD-Net. The qualitative results (Fig.~\ref{fig:FixQual}) highlighted the improved angular separation of fibres with low angular separation.  It is also evident that the overall shape of the FOD is captured, as opposed to discrete, or Dirac-like FODs \citep{elaldi2021equivariant,koppers2016direct,karimi2021learning}. Furthermore, statistically significant improvements were recorded in fixel accuracy, PAE and AFDE by $\text{SDNet}_{\kappa}$ across the white matter.

However, the introduction of fixel classification penalty in ROI-1-CC led to a reduction in fixel-based performance. This highlighted a potential bias of SDNet towards over-estimating the number of fixels in each voxel. The input of FOD reconstruction networks are necessarily derived from a DWI acquisition with low angular resolution, so do not have sufficient information to reconstruct FODs that contain all fixels, as observed in Fig.~\ref{fig:ERRmaps}. Therefore, the effect of the fixel classification penalty will generally be to correct these underestimations by encouraging the network to increase the number of fixels. Since ROI-1-CC contains only single fixel voxels, the fixel-classification penalty may have increased the number of over-estimations in this region, which, when combined with the already strong performance of SDNet and FOD-Net, led to the observed decrease in performance. On the other hand, in ROI-3-SLF, a region containing 3 crossing fibres, the use of fixel classification penalty improved performance compared to the other two deep learning methods, and despite worse performance in ROI-1-CC, $\text{SDNet}_{\kappa}$ resulted in an improvement in performance over the white matter voxels for all fixel-based performance metrics.

In the current work, the fixel classification network is trained on the ground truth data alone, which, depending on the efficacy of the FOD reconstruction algorithm, will have a different distribution to the reconstructed FODs. One possible approach to further improving performance is to devise an algorithm to jointly train the FOD reconstruction network and the fixel classification network, similar to the method used to train generative adversarial networks \citep{Goodfellow2014}.

 The fixel classification penalty component of the loss function appears to share some characteristics with regularisation terms that are ubiquitous in model-based methods for solving ill-posed inverse problems. In particular, to minimise a combination of SSE loss and the fixel classification penalty, a decrease in SSE was incurred, and we have identified in our validation experiments that the extent of such a sacrifice can be controlled by the adjustment of $\kappa$ (data not included). This suggests that the solution that obtains the lowest SSE may fail to capture certain desirable features of the FOD. In this work, we have highlighted this impact on the separation of fibre populations with similar orientations, but it is possible other features such as the continuity of fibre populations through space could also be improved using similar methods.

%% file: Sections/6.Conclusion.tex
In this work we have proposed SDNet, a model-based deep learning architecture optimised for FOD reconstruction. In addition to the learned regularisation blocks, are trained directly in an end-to-end fashion and therefore optimised for the task of FOD reconstruction, the network also takes a neighbourhood of multi-shell DWI signals as input to an architecture containing multiple cascades. We further show that there is a trade-off between FOD-based and fixel-based performance, and propose a fixel classification penalty term in our loss function, as implemented in $\text{SDNet}_{\kappa}$, as a method of controlling the the trade-off between these performance metrics. We show that, when compared to a state-of-the-art FOD super-resolution network, FOD-Net, gains in FOD-based and fixel-based performance were achieved by  SDNet and $\text{SDNet}_{\kappa}$, respectively.  

\section*{Acknowledgment}
We would like to thank Xi Jia from University of Birmingham for the fruitful discussion on network architecture and parameter tuning in this research. The computations described in this research were performed using the Baskerville Tier 2 HPC service (https://www.baskerville.ac.uk/). Baskerville was funded by the EPSRC and UKRI through the World Class Labs scheme (EP/T022221/1) and the Digital Research Infrastructure programme (EP/W032244/1) and is operated by Advanced Research Computing at the University of Birmingham.